\documentclass[11pt]{article}

\usepackage[final]{acl}

\usepackage{times}
\usepackage{latexsym}

\usepackage[T1]{fontenc}

\usepackage[utf8]{inputenc}

\usepackage{microtype}

\usepackage{inconsolata}

\usepackage{graphicx}

\usepackage{booktabs}
\usepackage{tabularx}
\usepackage{multirow}
\usepackage{amsmath}

\title{Do Social Patterns Hold in Synthetic Data? Analyzing Cyberbullying Dynamics in LLM-Generated and Authentic Dialogues}

\author{Arefeh Kazemi\textsuperscript{1} , Hamza Qadeer\textsuperscript{1} , Sinan Asci\textsuperscript{2} ,  \\ {\bf Joachim Wagner\textsuperscript{1} , Brian Davis\textsuperscript{1}} \\
\normalsize 
\textsuperscript{1} School of Computing, ADAPT Centre,
Dublin City University, Dublin, Ireland;\\ \textsuperscript{2} DCU Anti Bullying Centre;\\
 \small{\textsuperscript{1}\texttt{\{first.last\}@adaptcentre.ie} | \textsuperscript{2}\texttt{\{first.last\}@dcu.ie} }
}

\begin{document}
\maketitle
\begin{abstract}
Cyberbullying (CB) is a form of intentional and repeated online aggression characterized by a power imbalance between perpetrators and victims. Although large language models (LLMs) have recently been used to generate synthetic CB conversations for data augmentation and benchmarking, it remains unclear whether such data faithfully reproduces the underlying social dynamics of authentic interactions beyond supporting downstream task performance. In this paper, we present a comprehensive evaluation framework for assessing the social realism of LLM-generated CB conversations. We compare authentic and synthetic dialogues generated by GPT, Grok, and LLaMA across multiple complementary dimensions, including interactional structure (turn-taking, power dynamics, and repair behavior), linguistic and stylistic realism (pronoun usage and humor), affective and behavioral markers (cyberbullying types, profanity, and toxicity), and temporal escalation dynamics. We further complement automatic analyses with a human evaluation measuring CB presence, scenario relevance, role plausibility, and social realism. Our results show that LLM-generated data consistently preserves high-level interactional structures, including role participation patterns, directional power asymmetry between bullies and victims, and broad distributions of behavioral markers. However, all models systematically distort finer-grained social phenomena, including behavioral magnitude, role-specific allocation, categorical distributions, and temporal interaction dynamics. Moreover, these distortions are strongly model-dependent: GPT tends to suppress harmful content, Grok amplifies aggressive behaviors, and LLaMA provides the most balanced approximation while smoothing role distinctions. These findings demonstrate that synthetic CB data is a useful approximation of global conversational structure but remains an imperfect substitute for authentic interactions when behavioral realism and social dynamics are critical. Our work provides a socially grounded evaluation framework for synthetic dialogue and offers practical guidance for selecting and evaluating LLM-generated data in CB and other socially grounded NLP applications.
\end{abstract}

\section{Introduction}
Large language models (LLMs) have recently emerged as powerful tools for generating synthetic data across a wide range of natural language processing (NLP) tasks \cite{he-etal-2021-generate, he-etal-2022-generate, bonifacio-etal-2022-inpars, meng-etal-2022-generating}. By leveraging their ability to produce fluent and contextually coherent text, LLMs are increasingly used for data augmentation, few-shot learning, and synthetic dataset generation. In many cases, such synthetic data can substantially improve model performance and, under certain conditions, particularly when carefully filtered or combined with real data, can approach the performance of models trained on human-annotated datasets \cite{gupta2023targen, anaby-etal-2020-donot, li-etal-2023-synthetic}. In domains where labeled data is scarce or expensive to obtain, synthetic data generation offers a scalable and flexible alternative, enabling rapid development and evaluation of NLP systems.
One domain where these developments are particularly impactful is cyberbullying (CB) detection. Cyberbullying is a pervasive form of online aggression that disproportionately affects children and adolescents, involving intentional harm, repetition, and power imbalance \cite{hinduja2014bullying, patchin2006bullies}. Importantly, CB is inherently interactional and context-dependent, harmful intent often emerges through multi-turn exchanges involving bullies, victims, and bystanders, rather than from isolated messages \cite{sheth2022defining,ziems2020aggressive}. As a result, understanding and modeling CB requires capturing not only the content of individual messages but also the dynamics of conversation, including escalation patterns, role behaviors, and relational cues.
Despite its importance, collecting and annotating authentic CB data presents substantial challenges. Ethical and legal constraints limit access to conversations involving minors, particularly when they contain sensitive or harmful content \citeyear{facca2020exploring}. The creation of such datasets is constrained by high costs and the time-intensive nature of manual annotation, as well as ethical concerns, including exposure of annotators to distressing material and the risk of re-traumatization \cite{alemadi2024emotional}. In addition, adolescent communication is highly dynamic, shaped by evolving slang, multimodal expressions, and context-dependent meanings, making it difficult for static datasets to remain representative over time \cite{dembe2024impact,mcgillivray2022leveraging}. These challenges have created a persistent data bottleneck, limiting progress in context-aware CB detection and the computational modeling of peer aggression.
To address these limitations, recent work has explored the use of LLMs to generate synthetic CB data \cite{kazemi2025synthetic,kazemi2025synbullying}. Such datasets can incorporate multi-turn conversational structure, role-specific behaviors, and fine-grained annotations, while avoiding many of the ethical and legal constraints of real data collection. Empirical findings suggest that models trained on synthetic data can achieve performance close to those trained on authentic data for CB detection \cite{kazemi2025synthetic}. These results position synthetic data as a promising solution for scaling research in online safety.
However, an important question remains largely unaddressed: while synthetic data may support downstream task performance, does it faithfully reproduce the underlying social dynamics of real interactions? CB is not merely a lexical phenomenon but a complex social process involving patterns of dominance, resistance, escalation, and group behavior \cite{chandrasekharan2017you, cheng2017anyone, wulczyn2017ex, zhang2018conversations}. If synthetic dialogues fail to capture these deeper structures, then models trained on them may learn superficial signals while missing the mechanisms that characterize real-world bullying. This raises critical concerns about the validity of synthetic data for applications that require behavioral understanding, such as intervention design, policy development, and simulation of online communities.
Beyond evaluation, identifying systematic differences between synthetic and authentic data also provides an opportunity to improve generative models themselves. By uncovering where LLM-generated dialogues diverge from real interactions, whether in role behavior, escalation dynamics, or linguistic expression, we can better understand the limitations of current generation paradigms and design more effective prompting, alignment, or training strategies. In this sense, analyzing social realism is not only diagnostic but also prescriptive, offering a pathway toward generating synthetic data that more faithfully reflects human social behavior.
In this work, we address this gap by systematically evaluating whether LLM-generated CB dialogues preserve the social patterns observed in authentic conversations. Rather than focusing solely on task-based performance, we propose a multi-dimensional analytical framework grounded in computational social science and conversation analysis. Our framework examines three complementary dimensions: (i) interactional structure, including turn-taking patterns, power dynamics, and apology, defense; (ii) linguistic and stylistic realism, including pronoun usage, hedging, and emphatic devices; and (iii) affective and behavioral markers, including CB types, toxicity, profanity, sentiment, and escalation trajectories.
By comparing authentic and synthetic datasets across these dimensions, we assess the extent to which LLMs reproduce not only the language but also the social organization of CB interactions.
In addition to automatic analyses, we conduct a human evaluation to measure perceived realism along dimensions such as role plausibility, social coherence, and scenario consistency. This allows us to examine the relationship between subjective judgments of realism and objective structural properties, providing a more comprehensive understanding of the strengths and limitations of synthetic data.
This paper makes the following contributions: (1) we propose a socially grounded evaluation framework for analyzing cyberbullying conversations, capturing interactional, linguistic, and affective dimensions beyond surface-level text features.
(2) we conduct a comparative analysis between authentic and LLM-generated cyberbullying dialogues, as well as a cross-model comparison of multiple LLMs, evaluating how well different models reproduce these social dimensions. (3) we show that while synthetic data can approximate surface-level linguistic patterns, it fails to fully capture deeper social and interactional dynamics present in authentic conversations. (4) we include a human evaluation of conversational realism to complement automatic metrics.\\
The remainder of the paper is organized as follows. Section 2 reviews related work on cyberbullying detection, synthetic data generation, and evaluation of language models. Section 3 describes the datasets used in this study. Section 4 presents our analytical framework and metric design. Section 5 outlines the experimental setup. Section 6 reports the results of our quantitative analyses, followed by the human evaluation in Section 7. Section 8 discusses the implications of our findings, and Section 9 concludes the paper.
\section{Related Work}
\subsection{Cyberbullying Detection and Analysis}
CB has been widely studied as a form of online aggression characterized by intentional harm, repetition, and power imbalance \cite{patchin2006bullies,ejaz2024towards}. Early work in CB detection primarily focused on message-level classification, treating instances independently and relying on lexical and syntactic cues \cite{lopez2021early, dadvar2012improved}. While such approaches have achieved strong performance in toxicity detection, they often fail to capture the context-dependent and interactional nature of CB.
More recent studies emphasize that CB unfolds through multi-turn conversations, where meaning emerges through discourse dynamics, role interactions, and escalation patterns \citeyear{yi2023session, van2018automatic}. Fine-grained, role-aware datasets \cite{van2018automatic, sprugnoli-etal-2018-creating, ollagnier-etal-2022-cyberagressionado} have advanced CB research, but remain limited in scale, class balance, ecological validity, and youth representativeness. Furthermore, CB language evolves rapidly, incorporating slang, emojis, and implicit expressions \cite{dembe2024impact, mcgillivray2022leveraging}. Models trained on static data therefore struggle to generalize to emerging or obfuscated bullying patterns, such as sarcasm, exclusionary humor, or subtle peer manipulation. These limitations motivate the need for new data sources and evaluation frameworks that go beyond surface-level classification and better reflect the social dynamics of CB.

\subsection{Synthetic Data and LLM-Generated Text}
Recent advances in large language models (LLMs) have enabled scalable and high-quality synthetic data generation, providing controllable and often more ethically safe alternatives to manually annotated datasets. Given their strong capabilities in natural language understanding and generation, LLMs have been widely explored for synthetic data creation across a variety of NLP tasks. Early work leveraged LLMs for data augmentation, knowledge distillation, and few-shot learning \cite{anaby-etal-2020-donot, he-etal-2021-generate, he-etal-2022-generate, bonifacio-etal-2022-inpars, meng-etal-2022-generating, yoo-etal-2021-gpt3mix}, where synthetic samples are used to train or fine-tune downstream models, including sequence classifiers and retrieval systems. For instance, \newcite{he-etal-2021-generate, he-etal-2022-generate} explored LLM-generated data for knowledge distillation, self-training, and few-shot learning, with annotations produced by state-of-the-art classifiers. Similarly, \newcite{bonifacio-etal-2022-inpars} used LLMs to generate labeled data for information retrieval in a few-shot setting, which was then used to fine-tune smaller retrieval models for reranking. In addition, \newcite{yoo-etal-2021-gpt3mix} proposed augmenting training data by embedding task-specific sentences into prompts, while \newcite{anaby-etal-2020-donot} fine-tuned a language model on limited labeled data and subsequently used it to generate additional labeled samples. Moreover, \newcite{meng-etal-2022-generating} explored class-conditioned generation using label-descriptive prompts for classification tasks.
Beyond general NLP applications, synthetic data generation has also been explored in domain-specific settings. In the medical domain, \newcite{wang-etal-2024-notechat} introduced NoteChat, a multi-agent framework for generating synthetic patient–physician conversations from clinical documents. In the context of mental health, \newcite{ghanadian-etal-2024-socially} constructed socially aware synthetic datasets for suicidal ideation detection by identifying relevant social factors from psychological literature and generating data using zero-shot and few-shot prompting strategies. These studies demonstrate the adaptability of LLMs to sensitive and specialized domains. However, prior work has reported mixed findings regarding whether LLM-generated data can match the effectiveness of training on authentic datasets \cite{li-etal-2023-synthetic}.
In the domain of online harm detection, synthetic data has been applied to toxic language detection \cite{schmidhuber-kruschwitz-2024-llm}, as well as to the generation of biased and harmful content \cite{electronics13173431, hui-etal-2024-toxicraft}.
More recent studies emphasize that CB unfolds within multi-turn conversational structures, where meaning emerges through discourse dynamics, role interactions, and escalation patterns \cite{yi2023session, van2018automatic}. Recent work has also extended LLM-based synthetic generation to CB detection \cite{kazemi2025synthetic}. Among these, SynBullying \cite{kazemi2025synbullying} introduces a multi-LLM framework for generating synthetic conversational datasets specifically designed for CB detection, modeling interactional and role-based dialogue structures rather than isolated messages. By simulating multi-turn conversations among different participant roles, it provides a structured setting for studying CB in conversational contexts.
In a related direction, \cite{kazemi2025synthetic} investigates the effectiveness of LLM-generated synthetic data and labels compared to human-annotated (gold) data for CB detection. The study systematically evaluates when synthetic supervision can replace or complement gold-standard annotations, highlighting both the potential and limitations of relying on synthetic data for training CB classifiers.

\subsection{Beyond Accuracy: Evaluating LLM Behavior}

Traditional evaluation in NLP has largely relied on task-specific automatic metrics, such as accuracy, F1, BLEU, or ROUGE. While these metrics are effective for well-defined tasks with clear ground truth, they are insufficient for assessing the open-ended, generative, and socially situated behavior of large language models (LLMs). As LLMs increasingly generate long-form text, engage in dialogue, and simulate human-like interactions, evaluation has shifted toward multi-dimensional and behavior-oriented frameworks.
A growing body of work highlights the limitations of static benchmarks and aggregate scores. For example, \cite{li2023beyond} shows that evaluation based on fixed datasets fails to capture dynamic, interactive capabilities of LLMs, particularly in multi-turn or real-world scenarios. Similarly, analyses of LLM performance on human exams show that high scores do not necessarily reflect underlying reasoning ability or general competence, motivating more fine-grained evaluation approaches \cite{calais2024beyond}. These findings suggest that single-score evaluations can obscure important differences in model behavior.
To address these limitations, several works propose multi-dimensional evaluation frameworks that assess diverse aspects of model behavior. For instance, \cite{lin2023llm} introduces a unified, LLM-based framework for automatic evaluation of open-domain dialogue systems, assessing aspects such as coherence, relevance, and informativeness. Similarly, HELM (Holistic Evaluation of Language Models) \cite{liang2022holistic} proposes a holistic evaluation framework that extends beyond accuracy, incorporating multiple dimensions such as robustness, fairness, calibration, and efficiency. These approaches emphasize that LLM evaluation should capture multiple complementary aspects of performance rather than relying solely on predictive accuracy.
Another important line of work focuses on human evaluation and preference-based assessment. Because many aspects of text quality—such as fluency, helpfulness, and appropriateness—are inherently subjective, human judgments remain an important component of LLM evaluation. However, recent studies show that human evaluation is itself noisy and uncertain, and that simple correlation-based comparisons between automatic metrics and human judgments can be misleading \cite{elangovan2024beyond}. This has led to increased interest in combining human and model-based evaluators, including the use of LLMs as judges.
The paradigm of LLM-as-a-judge has emerged as a scalable alternative to human evaluation, where one model evaluates the outputs of another. While this approach can show reasonable agreement with human judgments in some settings, it also introduces challenges such as bias, self-preference, and evaluation instability. Recent work highlights the need to audit and improve LLM-based evaluators, as they can exhibit systematic reasoning and evaluation errors that may propagate into downstream assessments \cite{hasanbeig2023allure}.
Beyond text quality, researchers have increasingly begun to evaluate behavioral and social properties of LLMs. For example, multi-turn evaluation frameworks have been proposed to measure anthropomorphic behaviors such as empathy, role consistency, and relationship-building, which emerge over extended interactions \cite{ibrahim2025multi}. \cite{lu2025llm} examines the ability of LLMs to simulate human behavior using real-world data, highlighting the importance of evaluating objective behavioral fidelity in addition to perceived human-likeness. These studies emphasize that evaluating LLMs requires analyzing how models behave in interaction, not just the fluency of their outputs.
Finally, \cite{avraham2026dream} highlights fundamental challenges in evaluating generative models and motivates the use of agentic evaluation frameworks that assess deeper task performance beyond surface-level text quality. This is particularly relevant in socially grounded and multi-step tasks, where outputs may appear plausible while failing to capture underlying task requirements or interactional structure.
Overall, the literature demonstrates a clear shift from accuracy-centric evaluation toward holistic, multi-dimensional, and behavior-aware assessment of LLMs. However, despite these advances, there remains a lack of work examining whether LLM-generated data preserves social and interactional realism, especially in complex domains such as cyberbullying. Our work builds on this line of research by proposing a socially grounded, multi-dimensional evaluation framework and applying it to both authentic and synthetic conversational data, enabling a deeper analysis of how well LLMs capture real-world social behavior.

\section{Data}
\subsection{Authentic Dataset}\label{s:authentic-data}
To enable a comparative analysis between authentic and LLM-generated synthetic CB dialogues, we employ an existing CB dataset constructed through controlled teen role-play sessions \cite{sprugnoli-etal-2018-creating}. Specifically, we use the English version of this dataset introduced by \cite{verma-etal-2023-leveraging}.
In this dataset, each conversation is generated within a structured role-playing framework in which participants are assigned predefined roles that reflect real-world CB dynamics, namely: \textit{Victim}, \textit{Bully}, \textit{Bully Supporter}, and \textit{Victim Supporter}. This role-based design facilitates the emergence of socially meaningful interaction patterns and power asymmetries that are characteristic of CB scenarios.
All conversations are initiated using one of four carefully designed CB-triggering scenarios (labeled A–D), each representing a plausible situation that may lead to peer conflict and online harassment. Scenario A focuses on the gendered division of sports practices, portraying a shy male student who challenges gender norms by inviting peers to his ballet performance. Scenario B captures interference in others’ affairs, where a high-achieving student reports classmates for bringing cigarettes to school, resulting in disciplinary action and subsequent social exclusion. Scenario C reflects lack of independence and parental intervention, where increased homework is imposed following parental involvement, leading to resentment among students. Finally, Scenario D illustrates web virality, in which a shy student becomes the target of ridicule after an awkward dancing video circulates online.
The dataset is annotated by expert annotators using the fine-grained CB taxonomy introduced by \cite{van2018automatic}\footnote{While \newcite{van2018automatic} propose a broader annotation scheme, our work annotates only harmful messages with their corresponding cyberbullying (CB) types; consequently, the ``Defense'' category is excluded.}. The annotation scheme includes the following categories: \textit{Threat or Blackmail}, \textit{Insult General}, \textit{Insult Body Shame}, \textit{Insult Discrimination Sexism}, \textit{Insult Discrimination Racism}, \textit{Insult Attacking Relatives}, \textit{Curse or Exclusion}, \textit{Defamation}, \textit{Sexual Talk Harmless}, \textit{Sexual Talk Harassment}, \textit{Encouragement to Harassment}, and \textit{Other}.

\subsection{Synthetic Dataset}

To construct the synthetic counterpart of our analysis, we use "SynBullying", a multi-LLM synthetic conversational dataset introduced by \cite{kazemi2025synbullying}. This dataset is specifically designed to generate large-scale, role-based CB dialogues that approximate real-world interaction patterns while addressing the limitations of authentic data collection.
Synthetic conversations are generated using three large language models: GPT-4o (Feb-2025 version)~\cite{openai-2024-gpt}, Llama-3.3-70B-Instruct~\cite{meta-2024-llama33}, and Grok-2 (Feb-2025 version)~\cite{xai-2024-grok}. All generated data are subsequently annotated using GPT-4o (Sept-2025 version). The prompt engineering process follows an iterative refinement strategy: starting from an initial template, prompts are progressively improved based on qualitative evaluation of model outputs on a development set, with the aim of increasing the relevance, coherence, and consistency of both the generated conversations and their corresponding annotations.
The generation process targets multi-turn conversations that explicitly contain CB interactions. To this end, a role-based prompting framework is employed in which eleven fictional teenage participants are assigned predefined roles: one victim, two bullies, four victim supporters, and four bully supporters. This configuration is designed to simulate complex group interactions and social dynamics that are characteristic of CB scenarios. To encourage the generation of harmful content within a controlled and research-oriented setting, the task is framed as part of an academic study on CB detection. The model is provided with a predefined CB scenario and instructed to generate realistic conversations that may include profanity and aggressive language. In cases where models refuse to produce such content, the prompt is reissued until the desired number of conversations is obtained.
To ensure consistency with the authentic dataset, the same role-play scenarios (A–D) from \cite{sprugnoli-etal-2018-creating} are incorporated into the prompting process, guiding both the narrative structure and the interactions among participants. The resulting output consists of ordered sequences of messages exchanged among participants, where each message is explicitly associated with a role (e.g., Victim, Bully 1, Bully 2, Victim Supporter 1–4, Bully Supporter 1–4). Collectively, these messages form coherent conversations representing complete CB incidents.
Following data generation, all messages are automatically annotated using LLM-based labeling. Each message is assigned a binary label indicating whether it is harmful or harmless, and harmful messages are further annotated with one or more CB type labels. Given the context-dependent nature of CB, annotation is performed at the conversation level rather than on isolated messages, with each labeling instance processing the full conversation and returning labels for all messages. Based on preliminary evaluations, GPT-4o demonstrates the strongest performance among the tested models for CB-related labeling tasks and is therefore used for all synthetic annotations. Each message receives two types of labels: (1) a binary indicator (\textit{is\_harmful} = yes/no), and (2) when applicable, a set of CB-type labels aligned with the taxonomy used in the authentic dataset described in Section \ref{s:authentic-data}. Notably, a single harmful message may be associated with multiple CB categories, reflecting the nuanced and overlapping nature of CB behaviors. The quality of the LLM-generated labels is further validated by comparing them against human gold annotations, with the results confirming the high quality and reliability of the automatically assigned labels.

\subsection{Datasets Statistics}
Table~\ref{tab:dataset_stats} summarizes the distribution of conversations and messages across the authentic WhatsApp dataset (WA) and the synthetic (LLaMA, GPT and Grok) datasets for each scenario.

\begin{table}[ht]
\centering
\begin{tabular}{llrr}
\hline
\textbf{Dataset} & \textbf{Scenario} & \textbf{ Conv} & \textbf{Messages} \\
\hline
Authentic &  &  &  \\
\hline
WA & All & 10 & 2192 \\
WA & A & 4 & 1077 \\
WA & B & 2 & 574 \\
WA & C & 2 & 130 \\
WA & D & 2 & 411 \\
\hline
Synthetic &  &  &  \\
\hline
LLaMA & All & 40 & 3300 \\
LLaMA & A & 16 & 1290 \\
LLaMA & B & 8 & 625 \\
LLaMA & C & 8 & 695 \\
LLaMA & D & 8 & 690 \\
\hline
GPT & All & 40 & 4770 \\
GPT & A & 16 & 1542 \\
GPT & B & 8 & 922 \\
GPT & C & 8 & 781 \\
GPT & D & 8 & 1525 \\
\hline
Grok & All & 40 & 3960 \\
Grok & A & 16 & 1600 \\
Grok & B & 8 & 760 \\
Grok & C & 8 & 800 \\
Grok & D & 8 & 800 \\
\hline
\end{tabular}
\caption{Dataset statistics by scenario}
\label{tab:dataset_stats}
\end{table}

\section{A Metric-Based Framework for Analyzing Social Dynamics in CB Dialogues}
To address our research question—\textit{Do social patterns hold in synthetic CB data?},we operationalize a set of metrics that capture complementary dimensions of CB interactions. These metrics are designed to reflect \textbf{interactional structure}, \textbf{linguistic and stylistic realism}, and \textbf{affective and behavioral signals}, all of which are well-established components of social dynamics in online abuse. For each metric, we provide a definition, theoretical motivation and justification, and a precise computation procedure.

\subsection{Interactional Structure}

\subsubsection{Turn-taking Patterns}

Turn-taking patterns capture how conversational participation is distributed across roles. Drawing on foundational work in conversation analysis \cite{sacks1974simplest}, turn-taking reflects responsiveness, participation structure, and control of the conversational floor. In CB contexts, prior research suggests that aggressors may dominate interactions or disproportionately contribute to the discussions, and shape conversational trajectories through repeated hostile engagement\cite{cheng2017anyone, ribeiro2018characterizing, wulczyn2017ex}. We operationalize turn-taking as the role-based message distribution. For each conversation and each role $r$, we compute:
\[
P(r) = \frac{N_r}{N_{\text{total}}}
\]
where $N_r$ is the number of labeled messages produced by role $r$, and $N_{\text{total}}$ is the total number of messages in the conversation. This yields a normalized distribution over roles, representing the relative participation of each role.

\subsubsection{Power Dynamics}

Power dynamics reflect the presence of dominance, authority, and control in interaction  \cite{fairclough2013language, van1993principles}. In CB contexts, prior research shows that harmful interactions are often shaped by interactional asymmetries, social reinforcement, and aggressive conversational influence patterns \cite{cheng2017anyone, mathew2019spread, vidgen2021learning}. Linguistically, these are often realized through directives, commands, accusations, and other coercive or evaluative speech acts which signal attempts to control, intimidate, or socially position others \cite{bousfield2008impoliteness, culpeper2011impoliteness, searle1976classification}. We identify power-related speech acts using a dialogue act classification approach. Specifically, we employ an LLM-based classifier (GPT) to label each message with dialogue acts such as \textit{command}, \textit{accusation}, \textit{apology}, and \textit{defense}. 

For each role $r$, we compute the proportion of power-related acts as:
\[
P(\text{power} \mid r) = \frac{N_{\text{command}, r} + N_{\text{accusation}, r}}{N_{\text{labeled}, r}}
\]
where $N_{\text{command}, r}$ and $N_{\text{accusation}, r}$ denote the number of messages produced by role $r$ that are labeled as commands or accusations, respectively, and $N_r$ is the total number of messages for that role.

\subsubsection{Apology and Defensive Language}

Repair-related language, including apologies and defensive statements, reflects interactional strategies for conflict mitigation and conversational repair \cite{schegloff1977preference}. Prior work on online toxicity and conversational conflict suggests that participants may attempt to justify, deny, or de-escalate hostility through defensive or conciliatory responses as interactions evolve \cite{madhyastha2023study, zhang2018conversations}. These interactional behaviors are difficult to capture using purely lexical representations because their interpretation often depends on conversational context, pragmatic framing, and implied social meaning \cite{sap2020social}. Using the same dialogue act classification framework, we identify messages labeled as apology or defense. For each role $r$, we compute:
\[
P(\text{repair} \mid r) = \frac{N_{\text{apology}, r} + N_{\text{defense}, r}}{N_r}
\]
where $N_{\text{apology}, r}$ and $N_{\text{defense}, r}$ denote the number of messages containing apology or defensive acts.

\subsection{Linguistic and Stylistic Realism}

\subsubsection{Pronoun Usage}

The use of pronouns serves as an indicator of perspective-taking, self-focus, and social orientation in language \cite{chung2011psychological, tausczik2010psychological}. Previous studies have shown that aggressive or confrontational messages often contain higher usage of second-person pronouns (e.g. ``you''), reflecting direct targeting and interpersonal focus, while defensive or self-oriented responses may involve more first-person pronouns (e.g. ``I'') \cite{hancock2007lying, wulczyn2017ex}. For each role $r$, we compute the relative frequency of the pronoun categories at the token level using spaCy for tokenization \cite{honnibal2017spacy}:
\[
P(\text{pronoun type} \mid r) = \frac{N_{\text{pronoun tokens}, r}}{N_{\text{tokens}, r}}
\]
where \textit{the type} of pronoun includes first- and second-person pronouns, and $N_{\text{pronoun tokens}, r}$ is the number of tokens of the given type produced by role $r$.

\subsubsection{Humor}

Humor functions as a social strategy in CB interactions, often used for mocking, sarcasm, or group bonding among aggressors. Its distribution across roles provides insight into how social alignment and aggression are expressed.

We detect humorous messages using an LLM-based classifier (GPT). For each role $r$, humor frequency is computed as:
\[
P(\text{humor} \mid r) = \frac{N_{\text{humor}, r}}{N_r}
\]
where $N_{\text{humor}, r}$ is the number of messages labeled as humorous for that role.

\subsection{Affective and Behavioral Markers}

\subsubsection{Cyberbullying Types}

CB interactions can manifest through different types of harmful behavior, such as insults, threats, defamation, etc. The distribution of these categories reflects the diversity and structure of abusive behavior. We use the annotated CB-type labels provided in the original SynBullying dataset \cite{kazemi2025synbullying}. For each dataset, we compute the distribution:
\[
P(\text{CB type} = c) = \frac{N_c}{N_{\text{CB}}}
\]
where $N_c$ is the number of instances of category $c$, and $N_{\text{CB}}$ is the total number of CB-labeled messages.

\subsubsection{Profanity}

Profanity is a key marker of verbal aggression and hostility. We detect profanity using a curated lexicon with pattern matching that accounts for censored variants (e.g., ``f*k'', ``sh*t'') and elongated forms. After custom tokenization that preserves such variants, profanity is normalized per 100 tokens:
\[
\text{Profanity Rate}_r = \frac{N_{\text{profane tokens}, r}}{N_{\text{tokens}, r}} \times 100
\]
where $N_{\text{profane tokens}, r}$ is the number of profane tokens produced by role $r$.

\subsubsection{Toxicity}

Toxicity captures the degree of harmfulness or offensiveness in language. We use a pretrained toxicity classifier (ToxicBERT; \cite{unitary2020toxicbert}) to assign a score in $[0,1]$ to each message. We compute toxicity as the mean message-level toxicity per role:
\[
\text{Toxicity}_r = \frac{1}{N_r} \sum_{m \in r} \text{toxicity}(m)
\]

\subsubsection{Escalation Dynamics}

Escalation dynamics capture how toxicity evolves over the course of a conversation, reflecting interactional buildup, reinforcement, or de-escalation of aggression \cite{cheng2017anyone, zhang2018conversations}. Unlike static metrics, this measure models temporal structure of interaction \cite{madhyastha2023study, pavlopoulos2020toxicity}. We compute escalation dynamics as follows:

\begin{enumerate}
    \item For each conversation, let $N$ be the total number of messages.
    \item For each message in position $j$, compute its normalized position:
    \[
    \text{NormalizedPos}(j) = \frac{j}{N}
    \]
    yielding values in $(0,1]$.
    \item Partition the interval $(0,1]$ into $B$ equal-width bins (e.g., $B=20$, bins = (0–0.05, 0.05–0.10, …)).
    \item For each dataset $d$ and bin $b$, compute the mean toxicity($Toxi$):
    \[
    \text{MeanToxi}(d, b) = \frac{1}{|M(d,b)|} \sum_{m \in M(d,b)} \text{toxi}(m)
    \]
    where $M(d,b)$ is the set of messages in dataset $d$ whose normalized position fall within bin $b$.
\end{enumerate}

This procedure produces a trajectory of toxicity over normalized conversation time, enabling comparison of escalation patterns across datasets.

\medskip

Together, these metrics provide a comprehensive and multi-level operationalization of CB social dynamics, allowing us to systematically evaluate whether synthetic data preserves not only surface-level features, but also deeper interactional and behavioral patterns.


\section{Results}
\subsection{Interactional Structure}
\subsubsection{Turn-taking Patterns}

\begin{figure}[!ht]
\begin{center}
\includegraphics[width=\columnwidth]{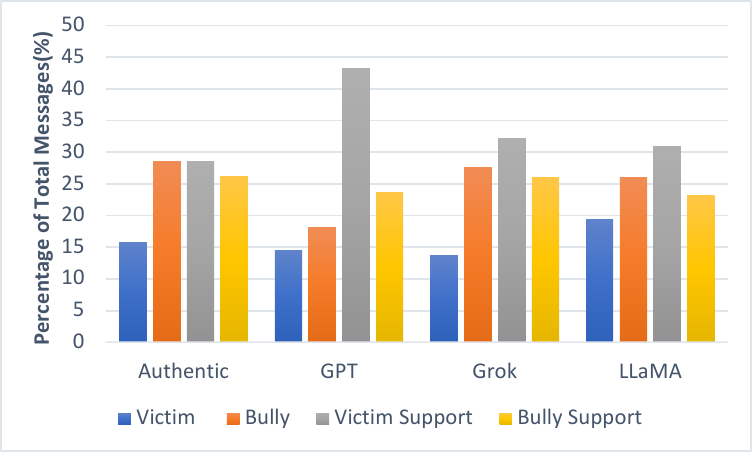}
\caption{Distribution of message contributions across conversation roles for authentic and synthetic datasets}
\label{f:turn-taking}
\end{center}
\end{figure}

Figure \ref{f:turn-taking} shows the distribution of message contributions by conversation roles (bully, victim, bully supporter, and victim supporter) across authentic and synthetic datasets. We assess distributional differences between these role-based message contributions using the chi-square test, reporting p-values alongside Cramér’s V and Jensen–Shannon divergence (JSD). Among the evaluated models, Grok most closely replicates the authentic role structure, exhibiting the smallest deviation (Cramér’s V = 0.042) and the lowest divergence (JSD = 0.0010), indicating an excellent match to the distribution of conversational roles. LLaMA also demonstrates strong alignment with the authentic data (Cramér’s V = 0.061; JSD = 0.0020), while GPT shows comparatively larger, but still modest deviations (Cramér’s V = 0.155; JSD = 0.0140).
Due to the large sample size, the chi-square test is highly sensitive and yields statistically significant results even for small deviations, therefore, p-values are interpreted solely as evidence of detectable differences rather than indicators of practical significance. 
Overall, the consistently low Cramér’s V and JSD values indicate that synthetic data closely approximates the authentic distribution of role participation across conversations. In particular, Grok and LLaMA demonstrate near-faithful reproduction of the authentic distribution of message contributions across roles, highlighting their strong ability to capture the overall conversational role structure.

\subsubsection{Power Dynamics}
\begin{figure}[!ht]
\begin{center}
\includegraphics[width=\columnwidth]{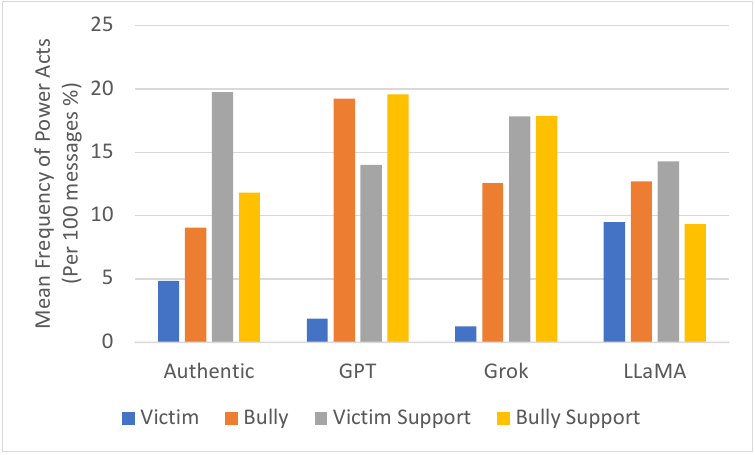}
\caption{Frequency of power acts across conversation roles for authentic and synthetic datasets}
\label{f:power-acts}
\end{center}
\end{figure}

Figure \ref{f:power-acts} presents the frequency of power-related acts across participant roles for each dataset. Across all datasets, a clear asymmetry between bully and victim roles is observed, with bullies consistently producing a higher proportion of power-related expressions than victims. This pattern is consistent with the established notion of power imbalance as a defining characteristic of cyberbullying interactions. In the authentic dataset, this difference is pronounced, with bullies exhibiting approximately twice the rate of power-related acts compared to victims (9.05\% vs. 4.84\%).

Synthetic datasets preserve this directional asymmetry but vary in the magnitude and distribution of role-specific behaviors. GPT and Grok exhibit larger bully–victim gaps than the authentic data, whereas LLaMA produces a comparatively smaller separation, with values that are closer in magnitude across roles.
At the role level, victim supporters constitute the most active group in the authentic dataset (19.75\%), a pattern that is only partially reproduced in synthetic data. GPT shifts this prominence toward bully supporters, making them the most active role (19.56\%), while Grok yields nearly identical levels for victim and bully supporters, indicating a balanced distribution between supporting roles but not across all roles. LLaMA, in turn, exhibits a more compressed and less differentiated role structure, including a comparatively elevated level of victim participation relative to other synthetic datasets.

Statistical comparisons indicate significant differences between authentic and synthetic distributions (all p < 0.001), although such results should be interpreted cautiously given the likely influence of large sample sizes. Effect size estimates (Cramér’s V $\approx$ 0.08–0.10) suggest small but consistent deviations, while low Jensen–Shannon divergence values $\approx$ 0.01–0.014) indicate that the overall distributional structure is broadly preserved across datasets.
Overall, the findings suggest that LLM-generated data consistently preserve the directional pattern of role-based power asymmetry in CB interactions. However, they differ from authentic data in the magnitude of this asymmetry and in the allocation of power-related behaviors across roles, particularly among supporting participants.

\subsubsection{Apology and Defensive Language}
\begin{figure}[!ht]
\begin{center}
\includegraphics[width=\columnwidth]{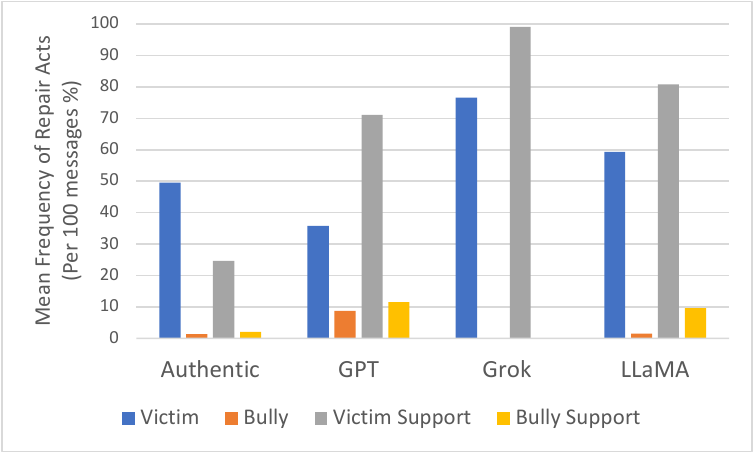}
\caption{Frequency of repair acts across conversation roles for authentic and synthetic datasets}
\label{f:repair-acts}
\end{center}
\end{figure}
Figure \ref{f:repair-acts} presents the frequency of repair-related markers (apologies and defensive language) across participant roles for each dataset. In the authentic dataset, repair behavior is strongly concentrated in victim and victim-support roles, with particularly high frequencies of defensive language (49.57\% for victims and 24.68\% for victim supporters), while bully and bully-support roles exhibit minimal repair activity (1.43\% and 2.08\%, respectively). This distribution reflects a clear role asymmetry in which targets of aggression and their supporters are the primary producers of repair-oriented language.
Across synthetic datasets, this general pattern (higher repair activity in victim-aligned roles than in bully) aligned roles—is partially preserved, but with substantial distortions in magnitude and distribution. GPT reproduces the directional pattern but substantially redistributes repair behavior, with elevated frequencies across all roles and especially high concentrations in victim-support interactions (71.01\%), indicating an overproduction of defensive language. Grok exhibits a more extreme deviation: repair acts are almost entirely absent in bully and bully-support roles (0\%), while being highly concentrated in victim and victim-support roles (76.6\% and 99.06\%, respectively), suggesting a collapse of repair behavior into a narrow subset of roles. In contrast, LLaMA shows a more balanced distribution, with repair activity present across all roles and a less extreme concentration in victim-aligned roles (59.29\% and 80.72\%), more closely approximating the relative structure observed in the authentic dataset, although still amplifying overall frequencies.
From a statistical perspective, differences between authentic and synthetic distributions are substantial. All models show significant deviations (p < 0.001), with Cramér’s V indicating moderate effects for GPT (0.197) and Grok (0.182), and a smaller but non-negligible effect for LLaMA (0.109). Jensen–Shannon divergence values further confirm these differences, with GPT (0.082) and Grok (0.062) showing pronounced divergence from the authentic distribution, and LLaMA exhibiting comparatively lower but still meaningful divergence (0.025). Compared with power dynamics, repair-related behaviors are reproduced with substantially lower fidelity across all models. Overall, while synthetic data retains the broad tendency for repair acts to be associated with victim-aligned roles, it diverges substantially from authentic data in the magnitude, distribution, and role-specific allocation of apology and defensive language.

\subsection{Linguistic and Stylistic Realism}
\subsubsection{Pronoun Usage}
Table~\ref{tab:pronoun_usage} presents the distribution of first- and second-person pronoun usage across different participant roles and datasets. For brevity, we use \textit{BS} to denote \textit{Bully Support} and \textit{VS} to denote \textit{Victim Support}.

\begin{table}[ht]
\centering
\begin{tabular}{llcc}
\hline
\textbf{Dataset} & \textbf{Role} & \textbf{1st (\%)} & \textbf{2nd (\%)} \\
\hline
Authentic & Bully   & 3.65  & 7.51 \\
Authentic & Victim  & 7.58  & 5.69 \\
Authentic & BS      & 4.04  & 6.43 \\
Authentic & VS      & 3.51  & 7.20 \\
\hline
GPT & Bully   & 3.96  & 3.61 \\
GPT & Victim  & 10.98 & 2.54 \\
GPT & BS      & 4.20  & 2.73 \\
GPT & VS      & 4.25  & 3.65 \\
\hline
Grok & Bully   & 1.91  & 4.93 \\
Grok & Victim  & 13.02 & 1.53 \\
Grok & BS      & 1.70  & 3.69 \\
Grok & VS      & 1.51  & 4.54 \\
\hline
LLaMA & Bully   & 3.16 & 5.99 \\
LLaMA & Victim  & 9.65 & 3.30 \\
LLaMA & BS      & 3.30 & 3.91 \\
LLaMA & VS      & 4.41 & 3.71 \\
\hline
\end{tabular}
\caption{Distribution of first- and second-person pronoun usage across roles and datasets}
\label{tab:pronoun_usage}
\end{table}

The authentic dataset exhibits a consistent but modest linguistic asymmetry aligned with prior observations in CB interactions: bully roles tend to use more second-person pronouns (e.g., “you”), commonly associated with direct address, while victim roles show higher use of first-person pronouns (e.g., “I”), often linked to self-reference and defensive positioning.
This role-conditioned pattern is broadly preserved in synthetic data, although with varying fidelity across models.
Among the evaluated models, LLaMA shows the closest alignment with the authentic distribution (Cramér’s V = 0.093; JSD = 0.0137), with no statistically reliable evidence of divergence at conventional thresholds (p = 0.086). GPT also maintains these patterns with only minor deviations (V = 0.092; JSD = 0.0156), although differences are statistically detectable (p = 0.033), indicating a modest redistribution of pronoun usage across roles. In contrast, Grok exhibits the largest divergence (V = 0.157; JSD = 0.0389; p < 0.001), reflecting a more substantial shift in role-conditioned pronoun usage.
Overall, these findings suggest that LLMs capture the general association between conversational roles and pronoun usage in CB contexts, although the strength and consistency of these patterns vary across models.

\subsubsection{Humor}

\begin{figure}[!ht]
\begin{center}
\includegraphics[width=\columnwidth]{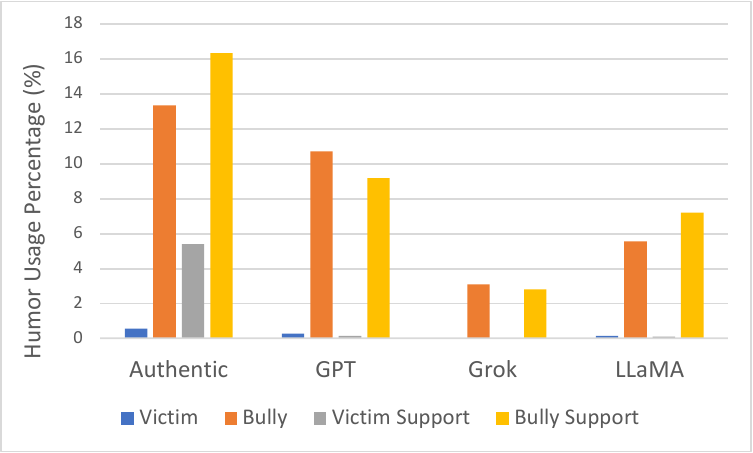}
\caption{Humor usage percentage across conversation roles for authentic and synthetic datasets}
\label{f:humor}
\end{center}
\end{figure}

Figure \ref{f:humor} presents the humor Usage Percentage by roles across datasets. In the authentic dataset, humor exhibits a clear asymmetric pattern: it is predominantly associated with bully-aligned roles (bully: 13.33\%, bully support: 16.32\%), while it is substantially lower in victim support (5.41\%) and nearly absent in victim messages (0.57\%). This indicates that humor is primarily concentrated in aggressor-aligned interactions, with limited presence in victim-aligned roles.
This role-conditioned pattern is only partially preserved in synthetic datasets, with notable deviations across models. GPT maintains higher humor usage in bully-aligned roles (bully: 10.7\%, bully support: 9.19\%); however, it substantially reduces humor in victim-support interactions (0.14\%) compared to the authentic data (5.41\%), resulting in a sharper separation between aggressor- and victim-aligned roles. Grok further distorts this pattern by drastically reducing overall humor usage (bully: 3.09\%, bully support: 2.8\%) and eliminating it entirely in victim and victim-support roles (0\%), thereby restricting humor to bully-aligned interactions. LLaMA shows a similar directional pattern, with higher humor usage in bully-aligned roles (bully: 5.56\%, bully support: 7.21\%) than in victim roles, but overall frequencies are attenuated relative to the authentic dataset, and humor remains nearly absent in victim-aligned roles (victim: 0.15\%, victim support: 0.10\%).
From a distributional perspective, all models diverge from the authentic data. Cramér’s V values indicate differences in the small-to-moderate range (GPT: 0.146; Grok: 0.126; LLaMA: 0.130), while Jensen–Shannon divergence highlights varying degrees of distributional shift. Grok exhibits the largest divergence (JSD = 0.0643), consistent with the absence of humor in victim-aligned roles, whereas GPT (0.0377) and LLaMA (0.0431) show moderate deviations. Although chi-square tests detect statistically significant differences across all models (p < 0.01), interpretation focuses on effect sizes and divergence, as humor is a relatively sparse feature and some role categories contain very low counts.
Overall, these findings indicate that while synthetic data preserves the general association between humor and bully-aligned roles, it only partially reproduces the role-specific distribution of humor observed in authentic conversations, particularly the presence of humor in victim-support interactions. This suggests that humor, as a context-sensitive and socially mediated behavior, is only partially reproduced by language models, with a tendency toward simplification and increased role polarization in synthetic dialogue.
\subsection{Affective and Behavioral Markers}

\subsubsection{Cyberbullying Types}

\begin{figure}[!ht]
\begin{center}
\includegraphics[width=\columnwidth]{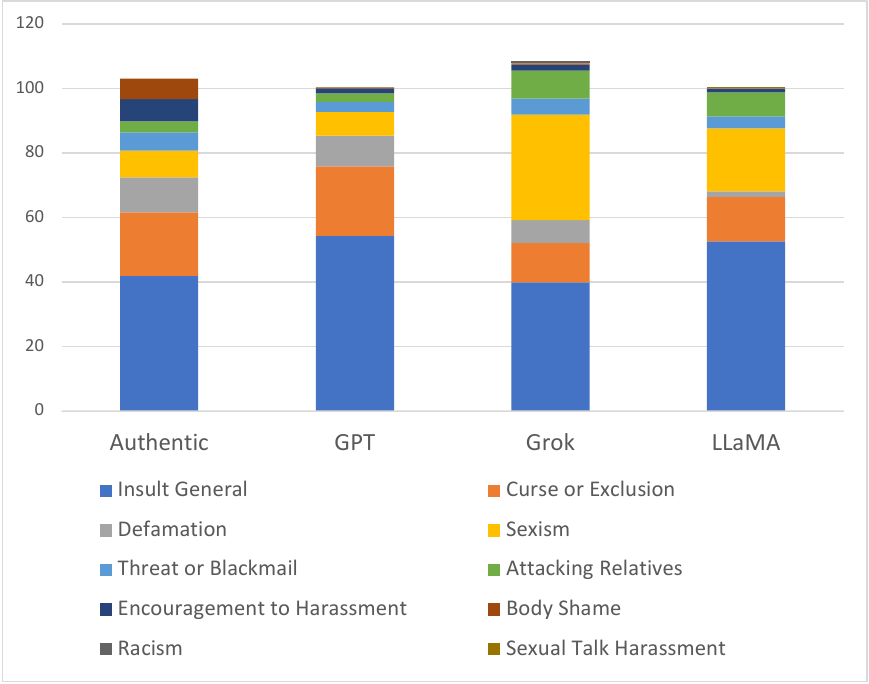}
\caption{Distribution of cyberbullying types across authentic and synthetic datasets}
\label{f:cb-types}
\end{center}
\end{figure}

\begin{table}[t]
\centering
\caption{Distributional similarity of cyberbullying types between authentic and synthetic datasets.}
\label{t:jsd_cbtypes}
\begin{tabular}{lcc}
\hline
Comparison & JSD & p-value \\
\hline
Authentic vs GPT   & 0.0317 & $1.11 \times 10^{-28}$ \\
Authentic vs Grok  & 0.0739 & $4.33 \times 10^{-68}$ \\
Authentic vs LLaMA & 0.0803 & $7.85 \times 10^{-59}$ \\
\hline
\end{tabular}
\end{table}

Figure \ref{f:cb-types} presents the distribution of CB types across datasets, while Table \ref{t:jsd_cbtypes} reports Jensen–Shannon divergence (JSD) and statistical significance for CB-type distributions. In the authentic dataset, Insult\_General is the dominant category (41.91\%), indicating that CB interactions are primarily expressed through generic insulting language rather than highly specific harm categories. Secondary CB types such as Curse\_or\_Exclusion (19.70\%), Defamation (10.79\%), and Insult\_Discrimination\_Sexism (8.41\%) also contribute substantially, reflecting a multi-faceted but highly skewed distribution of abusive behaviors.
This overall structure is partially preserved in synthetic datasets; however, notable distortions emerge in the distribution of secondary CB types. 
GPT most closely approximates the authentic CB-type distribution (JSD = 0.0317), preserving the dominance of Insult\_General and maintaining a broadly similar ranking of the major categories. However, GPT underrepresents several lower-frequency forms of cyberbullying, including Encouragement\_to\_Harassment and Insult\_Body\_Shame, resulting in a distribution that is more concentrated around general insults. In contrast, Grok exhibits a substantially larger divergence from the authentic distribution (JSD = 0.0739), primarily driven by a pronounced overrepresentation of Insult\_Discrimination\_Sexism and Insult\_Attacking\_Relatives, coupled with lower frequencies of Curse\_or\_Exclusion and Encouragement\_to\_Harassment. Grok is also the only model to generate a non-negligible proportion of Insult\_Discrimination\_Racism, which is absent in the authentic dataset. LLaMA shows the greatest divergence from the authentic distribution (JSD = 0.0803). Similar to Grok, it overrepresents Insult\_Discrimination\_Sexism and Insult\_Attacking\_Relatives, while markedly underrepresenting Defamation, Encouragement\_to\_Harassment, and Insult\_Body\_Shame. Despite these distortions, Insult\_General remains the most prevalent category. Although all synthetic distributions differ significantly from the authentic data (all $p < 10^{-28}$), GPT shows the closest alignment in overall distributional structure, while Grok and LLaMA exhibit stronger category-level distortions, particularly in identity-related insult categories.
Overall, these results suggest that while LLMs reliably reproduce the dominance of general insults in CB discourse, they struggle to preserve the fine-grained structure of CB-type diversity. In particular, synthetic data tends to redistribute rare and socially specific harm categories (e.g., discrimination-related insults) in model-dependent ways, indicating systematic biases in how different forms of abusive language are generated.

\subsubsection{Profanity}

\begin{figure}[!ht]
\begin{center}
\includegraphics[width=\columnwidth]{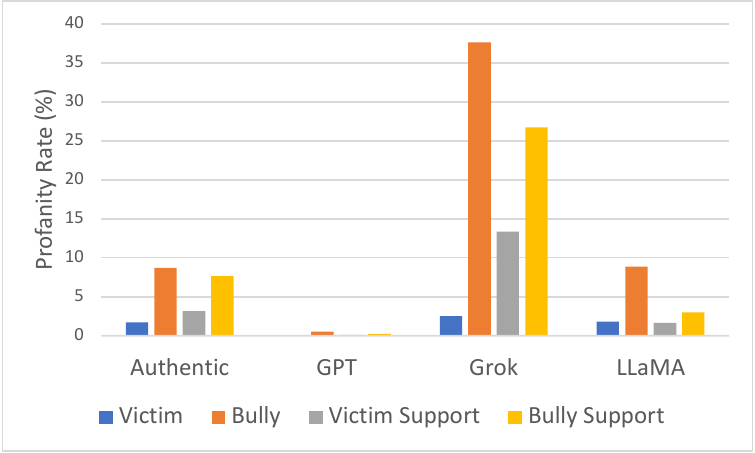}
\caption{Profanity rates across datasets and participant roles}
\label{f:profanity}
\end{center}
\end{figure}

Figure \ref{f:profanity} presents profanity rates across datasets and participant roles, while Table X reports Jensen–Shannon divergence (JSD) and statistical comparisons between authentic and synthetic datasets. 
In the authentic dataset, profanity shows a clear role-dependent pattern, with the highest usage observed in bully (8.73\%) and bully-support roles (7.64\%), followed by lower levels in victim-support (3.19\%) and minimal usage in victim messages (1.71\%). This aligns with established findings in CB research, where offensive and profane language is primarily produced by aggressors and reinforced through hostile interaction dynamics, while targets tend to use less abusive language \cite{cheng2017anyone, davidson2017automated, mathew2019spread, wulczyn2017ex}.
This pattern is partially preserved in synthetic datasets, but with important model-specific deviations. GPT exhibits a similar qualitative ordering across roles; however, it shows a dramatic reduction in overall profanity. This likely reflects the impact of strong safety and moderation filters in GPT-based systems, which suppress the generation of explicit profane or aggressive language. As a result, although the role structure is still visible, the magnitude of profanity is substantially attenuated.
In contrast, Grok amplifies profanity substantially across all roles, particularly in bully (37.64\%) and bully-support (26.74\%) messages. Despite this strong increase in magnitude, Grok preserves the relative ordering of roles observed in the authentic dataset (bully > bully-support > victim-support > victim), indicating that while the intensity of profanity is distorted, the underlying relational structure is largely maintained.
LLaMA shows a different pattern: profanity levels are closer in magnitude to the authentic dataset for the bully role (8.91\%), but the distribution is partially flattened across roles, with reduced separation between bully-support (3.01\%), victim (1.86\%), and victim-support (1.66\%). This suggests weaker role differentiation and a partial loss of the structure observed in authentic CB interactions.
From a distributional perspective, GPT shows the closest alignment with the authentic dataset (JSD = 0.0277; V = 0.062), while Grok (JSD = 0.0068; V = 0.051) preserves relative ordering but significantly distorts magnitude, and LLaMA exhibits moderate divergence (JSD = 0.0320; V = 0.145) with stronger redistribution across roles. 
Overall, these results indicate that the authentic dataset exhibits a stable role-based pattern in which profanity is concentrated among bully-aligned roles. Synthetic datasets partially reproduce this structure, but differ in systematic ways: GPT suppresses profanity due to safety alignment mechanisms, Grok amplifies profanity while preserving structural ordering, and LLaMA shows moderate redistribution across roles. Among the evaluated models, GPT shows the closest overall distributional alignment with the authentic data, although this similarity likely reflects suppression of profane language rather than faithful reproduction of authentic behavior. This suggests that profanity as a behavioral marker of CB is only partially and inconsistently captured in synthetic data, with model-specific biases affecting both its magnitude and role-based allocation.

\subsubsection{Toxicity}
\begin{figure}[!ht]
\begin{center}
\includegraphics[width=\columnwidth]{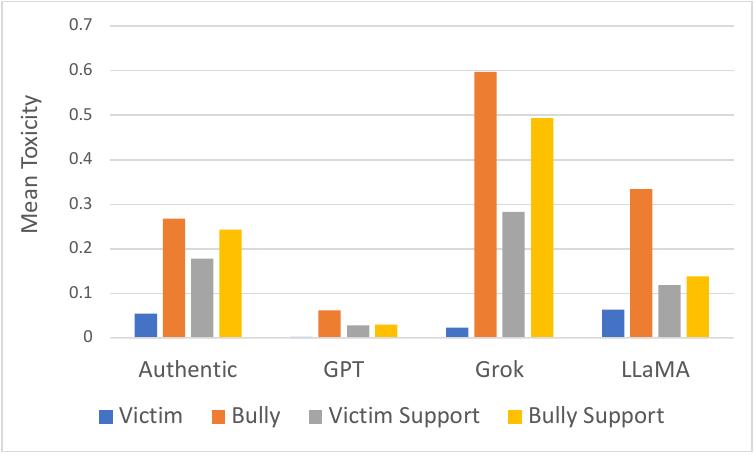}
\caption{Mean toxicity across datasets and participant roles}
\label{f:toxicity}
\end{center}
\end{figure}
Figure \ref{f:toxicity} presents the distribution of mean toxicity across participant roles
In the authentic data, toxicity exhibits a clear role-dependent structure, with the highest levels observed in the bully (0.268) and bully-support roles (0.243), followed by lower values in victim-support (0.178) and minimal toxicity in victim messages (0.055). This pattern reflects well-established dynamics in CB interactions, where harmful and aggressive language is predominantly produced by perpetrators and their allies, while targets exhibit substantially lower levels of toxicity \cite{cheng2017anyone, dinakar2011modeling, mathew2019spread, wulczyn2017ex, xu2012learning}.
This role-conditioned structure is largely preserved across synthetic datasets, as indicated by low divergence values for all models (JSD $\approx$ 0.012–0.016), suggesting that LLMs capture the relative distribution of toxicity across roles. In all models, bully-aligned roles exhibit higher toxicity than victim roles, consistent with the authentic data, although the strength of separation between roles varies.
However, substantial differences emerge in the magnitude of toxicity. GPT significantly underestimates toxicity across all roles, resulting in a relatively large mean absolute difference (0.1552), despite maintaining a similar distributional shape. In contrast, Grok substantially overestimates toxicity, leading to the largest deviation in magnitude (0.1790). Notably, Grok achieves the lowest JSD (0.0123), indicating that it closely matches the relative distribution of toxicity across roles while simultaneously exaggerating its overall intensity.
LLaMA provides the closest approximation to the authentic dataset in terms of overall magnitude (mean absolute difference = 0.0597), with toxicity values that are generally closer to the authentic levels. However, this approximation is accompanied by a redistribution across roles, including a marked reduction in bully-support toxicity relative to the authentic data and a narrower gap between roles overall. Its divergence remains low (JSD = 0.0156), indicating that it preserves the general structure of toxicity distribution, albeit with weaker role differentiation.
Overall, these findings indicate that LLMs consistently reproduce the relative structure of toxicity across roles, but differ substantially in how they model its intensity. GPT best preserves the relative ordering of roles but systematically underestimates toxicity, likely reflecting conservative generation behavior. Grok most accurately captures the distributional shape across roles (lowest JSD), but substantially overamplifies toxicity levels. LLaMA provides the closest match in terms of absolute magnitude, although it introduces distortions in the allocation of toxicity across roles. Taken together, these results suggest that no single LLM fully reproduces both the structure and scale of toxicity in authentic data, with each model capturing different aspects of the underlying behavioral pattern.

\subsubsection{Escalation Dynamics}
\begin{figure}[!ht]
\begin{center}
\includegraphics[width=\columnwidth]{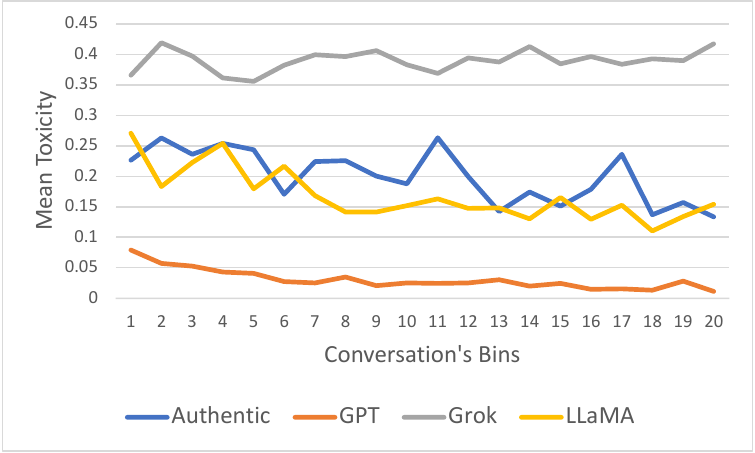}
\caption{Escalation curves of mean toxicity across conversation bins (1–20)}
\label{f:esclation}
\end{center}
\end{figure}
Figure \ref{f:esclation} illustrates the escalation curves of mean toxicity across conversation bins (1–20) for the authentic and synthetic datasets, while Table \ref{tab:toxicity_curve_similarity} reports quantitative comparisons between authentic and synthetic datasets using Jensen–Shannon divergence (JSD), L2 distance, Pearson correlation, and slope differences.
\begin{table}[t]
\centering
\small
\begin{tabular}{lcccc}
\toprule
\textbf{Model} & \textbf{JSD}$\downarrow$ & \textbf{L2}$\downarrow$ & \textbf{Pearson}$\uparrow$ & \textbf{Slope Diff.}$\downarrow$ \\
\midrule
GPT   & 0.0208 & 0.7754 & \textbf{0.5456} & 0.002844 \\
Grok  & 0.0067 & 0.8767 & $-$0.3294 & 0.006036 \\
LLaMA & \textbf{0.0062} & \textbf{0.2360} & 0.4934 & \textbf{0.000308} \\
\bottomrule
\end{tabular}
\caption{Comparison of toxicity escalation curves between authentic and synthetic datasets. Lower Jensen--Shannon divergence (JSD), L2 distance, and absolute slope difference indicate greater similarity to the authentic dataset, whereas higher Pearson correlation indicates stronger agreement in the overall shape of the toxicity escalation curve.}
\label{tab:toxicity_curve_similarity}
\end{table}

In the authentic data, toxicity exhibits a non-monotonic and multi-peaked temporal pattern, with higher values in early bins (0.226–0.263), mid-conversation fluctuations including a second peak around the middle segment (0.263), and a gradual decrease toward the final bins (0.157–0.134). This pattern suggests that cyberbullying interactions do not follow a simple linear escalation or decay process, but instead involve dynamic changes in aggression intensity across conversational stages, consistent with the interactional and response-driven nature of online aggression \cite{cheng2017anyone, madhyastha2023study, zhang2018conversations}.

Across synthetic datasets, all models partially reproduce the overall temporal variation of toxicity, but differ in their ability to preserve distributional similarity, temporal ordering, and absolute magnitude. GPT achieves the highest Pearson correlation with the authentic trajectory (r = 0.5456), suggesting that it most closely follows the overall direction of toxicity changes across conversation positions. However, it also exhibits the highest JSD (0.0208) and a relatively large L2 distance (0.7754), indicating that although the temporal ordering is relatively similar, the normalized toxicity distribution and absolute values differ from the authentic pattern.

Grok demonstrates a different behavior: it achieves a low JSD value (0.0067), indicating similarity in the normalized toxicity distribution, but shows a negative Pearson correlation (r = -0.3294), suggesting that its temporal toxicity trajectory follows an opposite direction compared with the authentic data. Its high L2 distance (0.8767) further indicates substantial differences in the magnitude of toxicity across conversation stages. This result highlights that distributional similarity alone does not guarantee preservation of conversational dynamics.

LLaMA provides the closest approximation in terms of absolute toxicity magnitude and overall temporal trend, achieving the lowest L2 distance (0.2360), a positive correlation with the authentic trajectory (r = 0.4934), and the smallest slope difference (-0.0003). These results suggest that LLaMA better preserves the global level and direction of toxicity evolution, although it may still differ from authentic conversations in finer-grained temporal variations.

Overall, the results demonstrate that synthetic models preserve different aspects of toxicity dynamics but do not consistently reproduce all properties simultaneously. GPT exhibits the strongest correspondence with the authentic trajectory in terms of temporal variation (highest Pearson correlation), Grok produces a similar normalized distribution but fails to preserve trajectory direction, and LLaMA most closely matches the magnitude and global trend of toxicity evolution. These findings indicate that reproducing conversational toxicity dynamics requires preserving not only the amount of toxicity expressed, but also when toxicity emerges and how it changes throughout an interaction.


\section{Human Evaluation}
While quantitative analyses capture distributional similarity and structural patterns, they do not fully assess whether synthetic conversations are perceived as realistic, contextually appropriate, and behaviorally plausible by humans. In particular, automatic metrics cannot determine whether interactions align with intended scenarios, exhibit credible role behavior, or are recognizable as cyberbullying. To address this limitation, we conduct a human evaluation focusing on cyberbullying validity, scenario coherence, role plausibility, and social realism.

\subsection{Evaluation Design}

The evaluation includes 9 authentic conversations \footnote{There are 10 authentic conversations in total; one of them was removed due to the length constraint. The removed conversation contains fewer than 10 messages.} and 24 synthetic conversations (2 per scenario A–D and per model: GPT, LLaMA, Grok), resulting in a total of 33 conversations. Each conversation is divided into three stages (early, middle, late), from which one 20-message snippet is sampled. This yields 99 snippets in total. Each snippet preserves turn order and includes role labels and scenario information, but does not reveal whether it is authentic or synthetic, ensuring unbiased judgments.

\subsection{Annotation Procedure}
Three annotators (one senior social science expert and two senior computer science experts) independently evaluated all snippets. Annotators were instructed to read each conversation snippet and assess it using the set of questionnaires presented in Table~\ref{tab:guidelines}. Annotators evaluated each snippet independently based solely on the provided context.

\begin{table*}[h]
\centering
\small
\begin{tabularx}{\textwidth}{p{3.2cm} X}
\toprule
\textbf{Question} & \textbf{Description} \\
\midrule
Cyberbullying Presence (Yes/No) & Does the snippet contain cyberbullying according to \cite{smith2008cyberbullying})'s definition: repeated, intentional aggression with power imbalance? \\
\midrule
Scenario Relevance (1–5) & “This snippet is relevant to the provided scenario.” \\
\midrule
Role Plausibility (1–5) & “The behaviors of the bully, victim, and supporters appear realistic for their roles.” \\
\midrule
Social Realism (1–5) & “The interaction feels like a plausible exchange between teenagers.” \\
\bottomrule
\end{tabularx}
\caption{Questionnaire used for human evaluation. Likert scale ranges from 1 (Strongly Disagree) to 5 (Strongly Agree).}
\label{tab:guidelines}
\end{table*}

\subsection{Inter-Annotator Agreement}

Agreement is measured using both Fleiss’ $\kappa$ and pooled multi-rater PABAK-OS (Table~\ref{tab:agreement}).

\begin{table}[h]
\centering
\small
\begin{tabular}{lcc}
\toprule
\textbf{Question} & \textbf{Fleiss’ $\kappa$} & \textbf{PABAK-OS} \\
\midrule
Cyberbullying Presence & 0.509 & 0.704 \\
Scenario Relevance & 0.012 & 0.917 \\
Role Plausibility & 0.073 & 0.787 \\
Social Realism & -0.030 & 0.861 \\
\bottomrule
\end{tabular}
\caption{Inter-annotator agreement across evaluation questions.}
\label{tab:agreement}
\end{table}
Fleiss’ $\kappa$ values are moderate for the binary task (0.509) but very low for Likert-scale questions. This discrepancy is expected due to the prevalence and bias problem of $\kappa$, which underestimates agreement when ratings are highly skewed (e.g., most responses concentrated in the 4–5 range). In contrast, PABAK-OS adjusts for prevalence effects and provides a more reliable estimate of agreement under such conditions. Given that annotators consistently assigned high ratings for scenario relevance and realism, $\kappa$ becomes artificially low despite strong consensus.
For this reason, PABAK-OS is more appropriate for interpreting agreement in our study. The high PABAK-OS values (0.787–0.917) indicate substantial to near-perfect agreement, particularly for Scenario Relevance (0.917) and Social Realism (0.861). Even for Role Plausibility (0.787), agreement remains strong. Overall, despite low $\kappa$ values, the consistently high PABAK-OS scores demonstrate that annotators show robust and reliable agreement, supporting the validity of the human evaluation.

\subsection{Human Evaluation Results}

Results are summarized in Table~\ref{tab:human_results}.

\begin{table}[h]
\centering
\small
\begin{tabular}{lccc}
\toprule
\textbf{Metric} & \textbf{Dataset} & \textbf{CB Frequency} & \\
\midrule
\multirow{4}{*}{CB Presence} 
& Authentic & 100\% &   \\
& GPT & 66.7\% &   \\
& Grok & 100\% &   \\
& LLaMA & 100\% &   \\
\midrule
\textbf{Metric} & \textbf{Dataset} & \textbf{Mean} & \textbf{Std} \\
\midrule
\multirow{4}{*}{Scenario Relevance}
& Authentic & 4.73 & 0.55  \\
& GPT & 4.57 & 0.63  \\
& Grok & 4.94 & 0.16  \\
& LLaMA & 4.74 & 0.47  \\
\midrule
\multirow{4}{*}{Role Plausibility}
& Authentic & 4.51 & 0.53  \\
& GPT & 3.94 & 0.81  \\
& Grok & 4.71 & 0.44  \\
& LLaMA & 4.40 & 0.58 \\
\midrule
\multirow{4}{*}{Social Realism}
& Authentic & 4.38 & 0.51  \\
& GPT & 4.11 & 0.53  \\
& Grok & 4.04 & 0.63  \\
& LLaMA & 4.36 & 0.56  \\
\bottomrule
\end{tabular}
\caption{Human evaluation results. CB presence is reported as majority vote percentage; other metrics are mean Likert scores with standard deviation.}
\label{tab:human_results}
\end{table}

\paragraph{Cyberbullying Presence.}
Authentic, Grok, and LLaMA datasets are consistently identified as containing CB (100\%). In contrast, GPT achieves only 66.7\%, indicating that a substantial portion of its outputs do not meet the CB definition. This confirms that GPT tends to suppress harmful content, reducing its suitability for CB data generation.
\paragraph{Scenario Relevance.}
All datasets achieve high scores (4.57–4.94), indicating strong alignment with assigned scenarios. Grok performs best (4.94), followed by LLaMA (4.74) and authentic data (4.73), while GPT is slightly lower (4.57). This suggests that LLMs reliably maintain scenario coherence.
\paragraph{Role Plausibility.}
Authentic data shows high plausibility (4.51). Grok achieves the highest score (4.71), indicating strong role differentiation, followed by LLaMA (4.40). GPT performs noticeably worse (3.94), suggesting weaker or less consistent role behavior.
\paragraph{Social Realism.}
Authentic conversations score 4.38. LLaMA closely matches this (4.36), indicating high realism. GPT (4.11) and Grok (4.04) are slightly lower, suggesting minor deviations in naturalness, potentially due to suppression (GPT) or amplification (Grok).

\section{Discussion}
This study addressed the central research question: \textit{Do social patterns hold in CB synthetic data?} The results provide a nuanced answer. \textbf{LLMs consistently reproduce several high-level social structures in CB interactions, but systematically distort their magnitude, distribution, and contextual nuance.} To elaborate on this finding, the discussion proceeds in three steps. We first examine the extent to which LLMs preserve core structural properties of CB interactions, including role-based participation patterns and power asymmetries. We then analyze where and how synthetic data diverges from authentic data, focusing on distortions in behavioral magnitude, role-specific allocation, and temporal dynamics. Finally, we consider model-specific biases and their broader implications for the use of synthetic data in NLP research.

\subsection{Preservation of Social Structure.}

Synthetic data consistently captures several core structural properties of CB interactions. In particular, LLMs reproduce \textbf{role-based turn-taking patterns}, the directional \textbf{power imbalance between bully and victim roles}, and the \textbf{relative ordering of behavioral markers such as toxicity and profanity} across participants. These patterns align with established findings in CB research, where harmful behavior is disproportionately produced by perpetrators and their allies, while targets exhibit comparatively lower levels of aggression \cite{cheng2017anyone, wulczyn2017ex, xu2012learning, zhang2018conversations}. The low divergence observed in these dimensions suggests that LLMs effectively learn and reproduce \textbf{coarse-grained statistical regularities} present in social CB interaction data, enabling them to approximate the overall structure of CB conversations\cite{aher2023using, argyle2023out, santurkar2023whose}.

Importantly, the human evaluation further supports this conclusion: synthetic conversations generated by Grok and LLaMA are consistently identified as containing cyberbullying (100\% CB presence), indicating that these models reliably reproduce the presence of abusive interactional structures. However, GPT shows a substantially lower CB presence rate (66.7\%), suggesting that a non-negligible proportion of its outputs do not contain clear CB behavior. This highlights that, while GPT preserves structural patterns at a distributional level, it is less reliable in consistently generating overt CB interactions.

\subsection{Distortions in Magnitude, Allocation, and Temporal Dynamics.}

Despite preserving high-level structure, LLMs exhibit systematic distortions in \textbf{behavioral magnitude, role-specific allocation, and temporal dynamics}. Across multiple metrics, models either amplify or suppress behaviors relative to authentic data, and redistribute them unevenly across roles. For example, harmful behaviors such as toxicity and profanity are often underproduced (e.g., GPT) or overamplified (e.g., Grok), while role-specific patterns, such as the distribution of power acts or repair behavior, are inconsistently allocated across participants.
These distortions are reinforced by human judgments of interaction quality. While scenario coherence remains high across all models, GPT receives lower scores in role plausibility (3.94 vs. 4.51 in authentic data), reflecting inconsistencies in how participant roles are enacted. Grok, although strong in role plausibility (4.71), tends to over-intensify aggressive behavior, which affects perceived naturalness. LLaMA achieves the most balanced performance, particularly in social realism (4.36 vs. 4.38 authentic), suggesting better preservation of overall interactional plausibility, but still exhibits role compression effects consistent with the automatic analyses.

In addition, LLMs struggle to reproduce \textbf{temporal interaction patterns}. While some models capture aspects of escalation, none simultaneously preserve the ordering, shape, and magnitude of behavioral trajectories over the course of a conversation. This indicates that, although LLMs capture \textit{who tends to do more than whom}, they do not reliably model \textit{how much} behavior is expressed or \textit{how it evolves over time}. These distortions likely arise because such properties depend on fine-grained interactional dynamics that extend beyond global distributional patterns.

\subsection{Model-Specific Biases.}

Deviations from authentic data are \textbf{systematic and model-dependent}, with each LLM exhibiting distinct strengths and weaknesses across different aspects of CB interaction modeling. Rather than a single model consistently outperforming others, the results indicate that model suitability depends on the specific dimension of CB behavior being analyzed.

\paragraph{GPT.}
A key limitation of GPT in this context is its tendency to underproduce or suppress CB content, likely due to strong safety and moderation mechanisms. Human evaluation confirms this limitation, showing that approximately \textbf{one-third of GPT-generated conversations do not contain identifiable CB (66.7\% CB presence)}, indicating incomplete reproduction of the target phenomenon. This suppression effect is reflected in consistently lower levels of toxicity, profanity, and other harmful behaviors compared to the authentic dataset.

Beyond this global attenuation, GPT also introduces distortions in role-specific allocation. For example, it overemphasizes certain roles in power dynamics (e.g., bully-support activity) and substantially redistributes repair behavior toward victim-support interactions, deviating from the authentic role structure. While GPT captures local temporal fluctuations in escalation patterns, it does not accurately reproduce their overall magnitude. Despite these limitations, GPT shows relatively strong distributional alignment with authentic data across several metrics, including power dynamics and CB-type distributions, and maintains consistent role-conditioned linguistic patterns such as pronoun usage. However, this apparent alignment is partly driven by suppression effects rather than faithful behavioral reproduction.

\paragraph{Grok.}
Grok shows a different profile, characterized by strong amplification of aggressive and identity-related behaviors. It produces the highest levels of toxicity and profanity, and substantially overrepresents discrimination-related CB types. In power dynamics, Grok exaggerates the bully–victim asymmetry, and in repair behavior, it collapses repair acts almost entirely into victim-aligned roles. Additionally, Grok fails to reproduce temporal interaction patterns, showing an inversion of escalation dynamics despite low divergence in normalized distributions. While Grok often preserves the directional ordering of roles (e.g., bully > victim in aggression), it does so with distorted magnitudes. Human evaluation further confirms this duality: Grok achieves high role plausibility and strong CB coverage (100\%), but this comes at the cost of intensified and less natural interactional dynamics.

Thus, Grok may be useful for capturing strong signal separation between roles, but is unreliable for realistic modeling of behavioral intensity, distributional balance, and temporal dynamics.

\paragraph{LLaMA.}
LLaMA exhibits a more balanced but smoothing-oriented behavior. Across multiple metrics, it produces values closer in magnitude to the authentic dataset compared to other models, particularly in toxicity and escalation dynamics, where it achieves strong alignment in overall trend and scale. It also shows strong alignment in linguistic patterns such as pronoun usage. Human evaluation further supports this pattern, with LLaMA achieving high CB presence (100\%), strong role plausibility (4.40), and near-authentic social realism (4.36).

However, LLaMA tends to redistribute behaviors across roles, resulting in weaker differentiation between participants. This is evident in power dynamics, where role distinctions are compressed, and in profanity and repair behavior, where separation between roles is reduced. Additionally, LLaMA introduces distortions in categorical distributions, particularly in CB types, through overrepresentation of certain categories. Overall, among the evaluated models, LLaMA provides the best trade-off between behavioral magnitude and temporal trends, but less effective in preserving sharp role distinctions and fine-grained distributional structure.

\paragraph{Summary.}
Table~\ref{tab:model_summary} summarizes the strengths, weaknesses, and recommended use of each model for generating synthetic CB data. Overall, the findings demonstrate that no single model fully captures all aspects of CB interactions. 



The differences among LLMs in generating synthetic CB data indicate that such data is inherently \textbf{model-dependent}, reflecting the inductive biases and alignment strategies of each model. Consequently, model selection should be guided by the specific analytical objective, rather than assuming uniform fidelity across behavioral dimensions.

\begin{table*}[t]
\centering
\small
\begin{tabular}{p{2cm} p{13.5cm}}
\toprule
\textbf{Model} & \textbf{GPT} \\
\midrule
\textbf{Strengths}
& Moderate alignment with global distributional structure, preservation of role-based conversational patterns, and consistent linguistic signals such as pronoun usage. Maintains relatively realistic CB-type distributions in several dimensions compared to other models. \\
\textbf{Weaknesses}
& Strongly constrained by safety and moderation mechanisms, resulting in systematic suppression of harmful content. In several cases, conversations contain weak or no explicit CB signals (66.7\% CB presence in human evaluation). Consequently, GPT shows consistently lower levels of toxicity and profanity and overproduces repair behavior, leading to distorted behavioral magnitude and inconsistent fidelity across CB dimensions. \\

\textbf{Recommended Use}
& Suitable for modeling global interaction structure and safe data generation; limited suitability for generating realistic harmful CB content due to systematic suppression of offensive language likely induced by alignment and safety filtering mechanisms. \\

\midrule

\textbf{Model} & \textbf{Grok} \\

\midrule
\textbf{Strengths}
& Preserves directional role-based patterns (e.g., bully $>$ victim) and produces strong, consistent CB signals (100\% CB presence). Maintains clear separation between roles and achieves low divergence in some normalized distributions. \\

\textbf{Weaknesses}
& Amplifies harmful behaviors beyond realistic levels, overrepresents discrimination-related insults, and distorts behavioral balance across roles. Fails to reproduce temporal dynamics and collapses some behaviors into limited roles despite low divergence in certain normalized distributions. \\

\textbf{Recommended Use}
& Suitable for tasks requiring clear signal separation and detection of CB patterns; less suitable for modeling realistic behavioral magnitude, balanced distributions, or temporal dynamics. \\

\midrule

\textbf{Model} & \textbf{LLaMA} \\

\midrule
\textbf{Strengths}
& Provides the best overall trade-off across behavioral magnitude, temporal dynamics, and linguistic realism. Achieves strong alignment in toxicity magnitude and escalation trends, and maintains high CB presence (100\%). \\

\textbf{Weaknesses}
& Smooths role distinctions, weakening role-specific allocation of behaviors. Distorts categorical distributions and produces less pronounced asymmetries in power and aggression. \\

\textbf{Recommended Use}
& Most suitable model for generating synthetic CB data when both behavioral realism and temporal structure are required; however, still limited in preserving sharp role differentiation and fine-grained categorical fidelity. \\

\bottomrule
\end{tabular}
\caption{Model-wise breakdown of strengths, weaknesses, and recommended usage for synthetic CB data generation.}
\label{tab:model_summary}
\end{table*}

\subsection{Implications for NLP Research.}

These findings highlight a nuanced role for synthetic CB data in NLP research and downstream applications. On the one hand, synthetic data is particularly useful in settings where the goal is to learn or model \textbf{coarse-grained interactional CB structure}. This includes capturing role-based asymmetries (e.g., bully versus victim dynamics), general conversational flow, and broad distributions of harmful versus non-harmful behavior. In such cases, synthetic data can serve as a scalable resource for pretraining, data augmentation, or exploratory analysis, especially when annotated real-world CB data is scarce, sensitive, or difficult to obtain.

On the other hand, the results show that synthetic CB data is \textbf{unreliable for tasks that depend on fine-grained social and pragmatic signals}. This includes modeling the intensity of harmful behavior, the distribution of rare but socially important categories (e.g., discrimination-related or context-specific insults), and interactionally grounded behaviors such as repair, humor, and escalation over time. In these settings, synthetic data introduces systematic biases, including over-smoothing of distributions, misallocation across participant roles, and distortions in temporal dynamics.

Crucially, the human evaluation shows that even when synthetic data appears structurally similar in quantitative metrics, it may still diverge in perceived realism or behavioral validity. GPT, in particular, demonstrates this mismatch: despite partial distributional alignment, a substantial fraction of its outputs do not contain recognizable CB interactions. Therefore, synthetic data should not be treated as a faithful proxy for real-world CB behavior in studies aiming to understand social mechanisms or to evaluate models in sensitive applications.

Overall, synthetic CB data should be viewed as a \textbf{structurally informative but behaviorally limited approximation} of real-world CB interactions: useful for capturing global patterns, but insufficient for faithfully representing nuanced social and pragmatic phenomena.


\section{Conclusion}
This paper investigated the extent to which social patterns of CB interactions are preserved in LLM-generated synthetic data. Through a comprehensive multi-dimensional analysis—covering interactional structure, linguistic behavior, affective signals, and temporal dynamics, we compared synthetic CB conversations generated by multiple LLMs with authentic CB data.
The findings provide a clear and nuanced answer to our central research question. Synthetic data reliably captures high-level structural properties of CB interactions, including role-based participation patterns, directional power asymmetry between bullies and victims, and broad distributions of behavioral markers such as toxicity, profanity, and pronoun usage. These results suggest that LLMs effectively learn and reproduce coarse-grained statistical regularities of social interaction. However, this structural fidelity does not extend to finer-grained behavioral dynamics. Across multiple dimensions, synthetic data exhibits systematic distortions in magnitude, role-specific allocation, and temporal evolution. Models either suppress or amplify harmful behaviors, misallocate behaviors across participant roles, and fail to accurately reproduce interactional trajectories such as escalation patterns. These limitations are particularly evident in context-sensitive phenomena such as repair behavior, humor, and the distribution of specific CB types.

Importantly, these deviations are not random but model-dependent, reflecting the underlying inductive biases and alignment strategies of each LLM. GPT tends to suppress harmful content, sometimes producing interactions with limited or no explicit CB, which reduces its suitability for generating realistic CB data. Grok amplifies aggressive and identity-related behaviors, preserving directional patterns but distorting intensity and category distributions. LLaMA provides the most balanced approximation overall, better capturing behavioral magnitude and temporal trends, although it smooths role distinctions and redistributes behaviors across participants.

These findings have important implications for the use of synthetic data in NLP research. Synthetic CB data is well-suited for tasks that rely on global structural patterns, such as modeling role dynamics or training systems on general interactional structure. However, it is less reliable for tasks requiring fine-grained social realism, including the study of behavioral intensity, rare or sensitive categories, and temporally grounded interaction dynamics. As a result, synthetic data should not be treated as a direct substitute for authentic data in applications that require high ecological validity.

Overall, this work demonstrates that while LLM-generated data offers a scalable and valuable resource, it remains a structurally informative but behaviorally imperfect proxy for real-world CB interactions. Future research should focus on improving the alignment of synthetic data with authentic social dynamics, particularly in terms of temporal modeling, role-specific behavior, and context-sensitive language use, as well as developing evaluation frameworks that integrate both computational and human-centered perspectives.

\appendix





\bibliography{custom}




\end{document}